\documentclass[runningheads]{llncs}
\usepackage[T1]{fontenc}
\usepackage{amsmath,amssymb,amsfonts}
\usepackage{graphicx}
\usepackage{xcolor}
\usepackage{threeparttable}
\usepackage{multirow}
\usepackage{booktabs}
\usepackage{colortbl}
\usepackage{array}
\usepackage{pifont}
\usepackage{bm}
\usepackage{placeins}

\newcommand{\boldres}[1]{{\textbf{\textcolor{red}{#1}}}}
\newcommand{\secondres}[1]{{\underline{\textcolor{blue}{#1}}}}

\definecolor{mycolor}{RGB}{241,240,255}

\def\BibTeX{{\rm B\kern-.05em{\sc i\kern-.025em b}\kern-.08em
    T\kern-.1667em\lower.7ex\hbox{E}\kern-.125emX}}
\begin{document}

\title{
DST-Net: Decomposed Spatio-Temporal Network for Time Series Forecasting
}
\titlerunning{DST-Net for Time Series Forecasting}


\author{
Yaohua Zha\inst{1,2} \and
Hai Wu\inst{1}\textsuperscript{\normalfont\ensuremath{\star}} \and
Shu-Tao Xia\inst{1,2}\thanks{Corresponding authors.}
}

\authorrunning{Yaohua Zha et al.}

\institute{
Pengcheng Laboratory \and
Tsinghua Shenzhen International Graduate School, Tsinghua University\\
\email{zyh1614399882@gmail.com, wuh05@pcl.ac.cn, xiast@sz.tsinghua.edu.cn}
}

\maketitle

\begin{abstract}
Series decomposition has emerged as a widely acknowledged paradigm in deep time series forecasting models. However, existing approaches predominantly employ a homogeneous modeling strategy for all subsequences, thereby overlooking the distinct inductive biases inherent in different components. We identify that the trend component exhibits dominant spatial coupling across channels but carries significant temporal redundancy, whereas the seasonal component is characterized by local temporal evolution but remains susceptible to attention distraction. To address this, we propose a novel disentangled architecture that tailors specific modeling strategies to align with the intrinsic properties of each component. For the trend branch, we design a Latent Graph Block, employing latent graph modeling to capture global topological dependencies and system-wide co-movements. Conversely, for the seasonal branch, we introduce a Gaussian Multi-scale Attention mechanism to mitigate attention distraction. By injecting learnable Gaussian priors, we impose structural constraints that compel the model to focus on multi-granular local contexts, effectively filtering out global noise and accurately capturing local evolutionary phases. Extensive experiments on multiple real-world benchmarks demonstrate the effectiveness of our proposed model.

\end{abstract}

\section{Introduction}
Time series forecasting plays a pivotal role in numerous real-world applications, such as weather forecasting \cite{weather1,weather2}  and economic analysis \cite{financial1,financial2}. However, real-world time series data typically exhibit extreme complexity characterized by the entanglement of non-stationarity and multi-scale variations. To tackle these intricate temporal patterns, the research community primarily relies on designing various sophisticated deep learning architectures, such as models based on Transformers \cite{autoformer,informer,itransformer,cploss} or MLPs \cite{amd,dlinear,timemixer}, which aim to capture latent complex relationships within sequences through powerful representation learning capabilities.

To alleviate modeling difficulty and decouple complex temporal variations, series decomposition has become a widely acknowledged paradigm in deep forecasting models. Since Autoformer \cite{autoformer} introduced decomposition as a fundamental block into deep networks, dismantling the original sequence into more predictable seasonal and trend components has proven effective. Subsequent studies, such as TimeMixer \cite{timemixer}, have further augmented the expressive power of decomposition architectures by utilizing multi-scale mixing techniques. Nevertheless, while existing methods have introduced numerous innovations regarding decomposition or fusion strategies, they frequently overlook the importance of differentiating the modeling of decomposed subsequences. Most existing methods employ identical network architectures (e.g., the same attention mechanism) to extract features from both high-frequency seasonal and low-frequency trend components. We argue that this uniform modeling approach fails to adequately account for the distinct properties of the decomposed components, thereby limiting the representation capability of the model.

\begin{figure}[t]
    \begin{center}
    \includegraphics[width=0.6\linewidth]{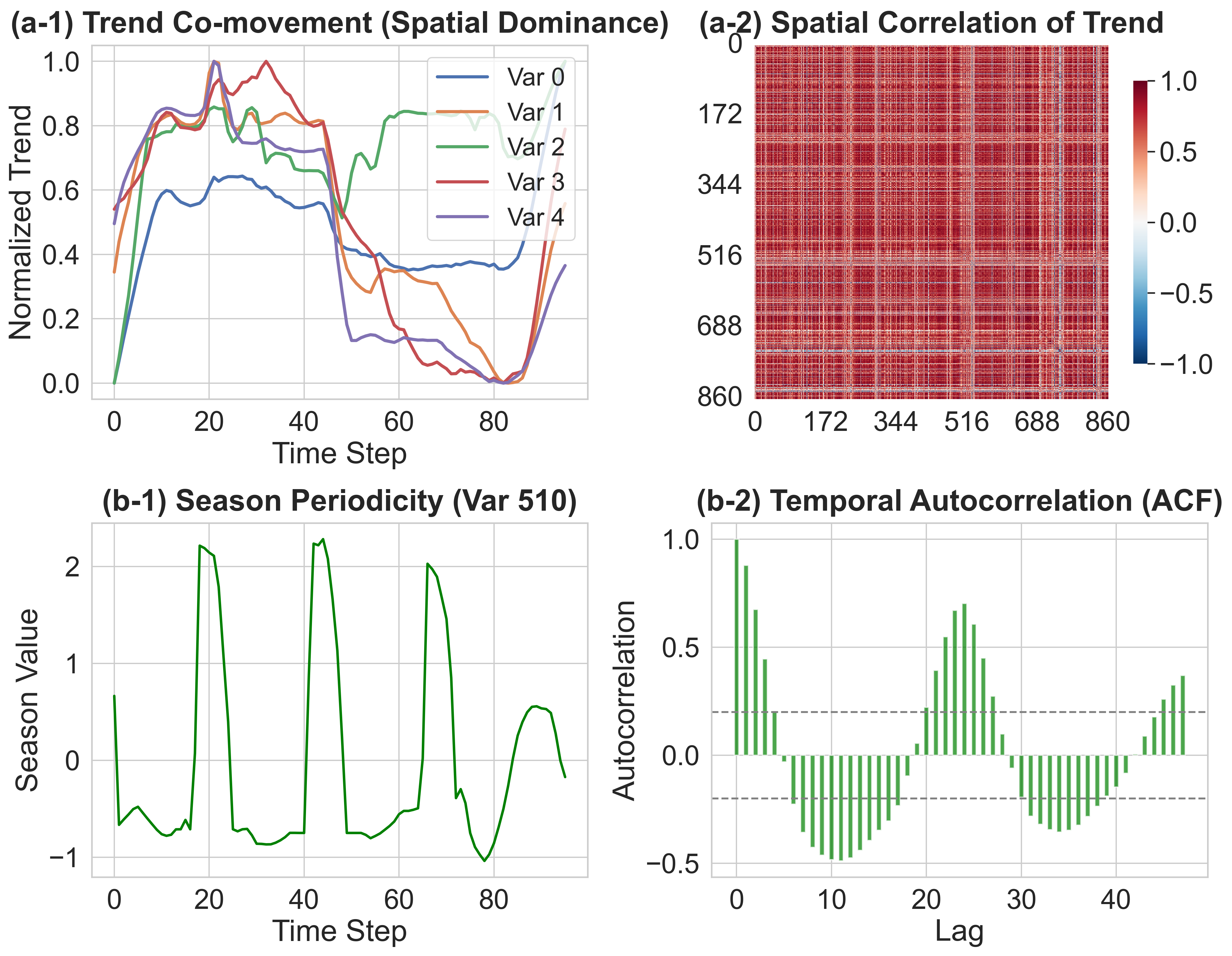}
    \caption{Statistical visualization of the decomposed components in Traffic dataset. (a) The trend component exhibits significant spatial co-movement and correlation among variates. (b) The seasonal component displays clear temporal periodicity and autocorrelation.
    }\label{motivation}
    \end{center}
\end{figure}

In this study, we delve into the intrinsic properties of the decomposed components and identify that seasonal and trend patterns exhibit distinct inductive biases. As illustrated in Fig. \ref{motivation}(a-1), distinct variables exhibit synchronized trend behavior, which is often determined by the underlying global state of the system, leading to strong coupling effects among variables. For instance, the onset of daily traffic peak hours acts as a latent global state that simultaneously drives up traffic occupancy rates across numerous upstream and downstream sensors, causing a synchronized rise in congestion levels throughout the entire network. This observation is quantitatively supported by the spatial correlation matrix in Fig. \ref{motivation}(a-2) where the dominance of high correlation values indicates that variables are deeply intertwined. Therefore, relying solely on temporal information is insufficient for trend modeling and necessitates the introduction of spatial dependency among variables to capture such system-level synergetic changes.

Conversely, the seasonal component is temporal-dominant and characterized by high-frequency fluctuations and periodic patterns. As illustrated in Fig.~\ref{motivation}(b-1), the evolution of the seasonal component exhibits distinct local continuity and periodicity, where the current state is strictly constrained by its immediate historical context and the evolutionary slope of the waveform. To further substantiate the temporal dependence of the seasonal component, we employed the Autocorrelation Function (ACF) analysis as shown in Fig.~\ref{motivation}(b-2). This plot was generated by calculating the correlation coefficients between the seasonal time series and its own lagged versions at various time steps. The results display significant positive peaks at periodic intervals (specifically at Lag 24), which quantitatively confirm that the seasonal component possesses strong self-correlation properties. This evidence reinforces the observation that the seasonal pattern is primarily driven by temporal self-evolution rather than external spatial factors, thereby justifying the focus on temporal modeling for the seasonal component.

Based on the aforementioned analysis, we propose a novel disentangled architecture that tailors distinct modeling strategies for the trend and seasonal components to align with their respective inductive biases. For the trend branch, we design a Latent Graph Block, employing latent graph modeling to capture global topological dependencies and system-wide co-movements. Conversely, for the seasonal branch, to address the distraction issue arising from distant numerical similarities, we introduce a Gaussian Multi-scale Attention mechanism. By injecting learnable Gaussian priors into the attention heads, we enforce a structural constraint that compels the model to focus on multi-granular local contexts, ranging from narrow views targeting high-frequency jitters to wider views capturing local periodic patterns. This design effectively filters out global noise and ensures the model captures the precise local evolution and phase shifts inherent in seasonal data. Finally, the temporal representations modeled independently in each branch are fused to yield the final predicted sequence. Extensive experiments on downstream datasets demonstrate the effectiveness of our proposed model.

\section{Related Work}

\subsection{Deep Time Series Forecasting Paradigms}

With the progressive breakthroughs in deep learning, the landscape of time series forecasting has evolved from RNN- and CNN-based models \cite{salinas2020deepar,bai2018empirical} to sophisticated Transformer variants. Early Transformer-based methods, such as Informer \cite{informer} and Autoformer \cite{autoformer}, adopted a Channel Mixing strategy, projecting multivariate features into a unified embedding space to capture latent correlations. However, as noted in recent studies \cite{NonstationaryTR,pyraformer}, this strategy often results in overfitting due to the entanglement of noise and signals, leading to inferior performance compared to simple linear baselines (e.g., DLinear \cite{dlinear}). Consequently, the paradigm shifted towards Channel Independence (CI), represented by PatchTST \cite{patchtst} and TiDE \cite{tilde}, which treats each variate independently and shares weights across channels. While CI significantly enhances robustness and data efficiency, it inherently sacrifices the capability to model cross-variable dependencies, creating a bottleneck in scenarios where system-wide coupling is critical. Hybrid approaches \cite{itransformer,crossformer} have attempted to revive Channel Dependence (CD) by renovating attention mechanisms to capture cross-variate interactions explicitly. Recent Mamba-based~\cite{mamba,lcm,pma} forecasters adapt selective state-space models to capture temporal and cross-variate dependencies. MambaTS~\cite{mambats} learns inter-variable dependencies to determine variable-aware scan orders, while TimePro~\cite{timepro} constructs variable- and time-aware hyper-states to preserve fine-grained temporal information during cross-variate modeling. Overall, these methods primarily model temporal and channel dependencies within a unified framework for the sequence as a whole, whereas DST-Net tailors its modeling strategies to the distinct characteristics of decomposed components to better capture temporal dynamics and cross-variate dependencies.

\section{Methodology}
\label{sec:methodology}

\subsection{Problem Definition and Overall Architecture}

\textbf{Problem Definition.} 
Let $\mathbf{X} = \{ \mathbf{x}_1, \dots, \mathbf{x}_L \} \in \mathbb{R}^{L \times N}$ denote the historical observed sequence, where $L$ is the lookback window size and $N$ is the number of variates. The objective of multivariate time series forecasting is to predict the future sequence $\mathbf{Y} = \{ \mathbf{x}_{L+1}, \dots, \mathbf{x}_{L+T} \} \in \mathbb{R}^{T \times N}$, where $T$ is the forecasting horizon. 
The core challenge lies in disentangling the complex, coupled spatiotemporal dependencies hidden within the non-stationary observations to achieve accurate predictions.

\begin{figure}[t]
    \centering
    \includegraphics[width=\linewidth]{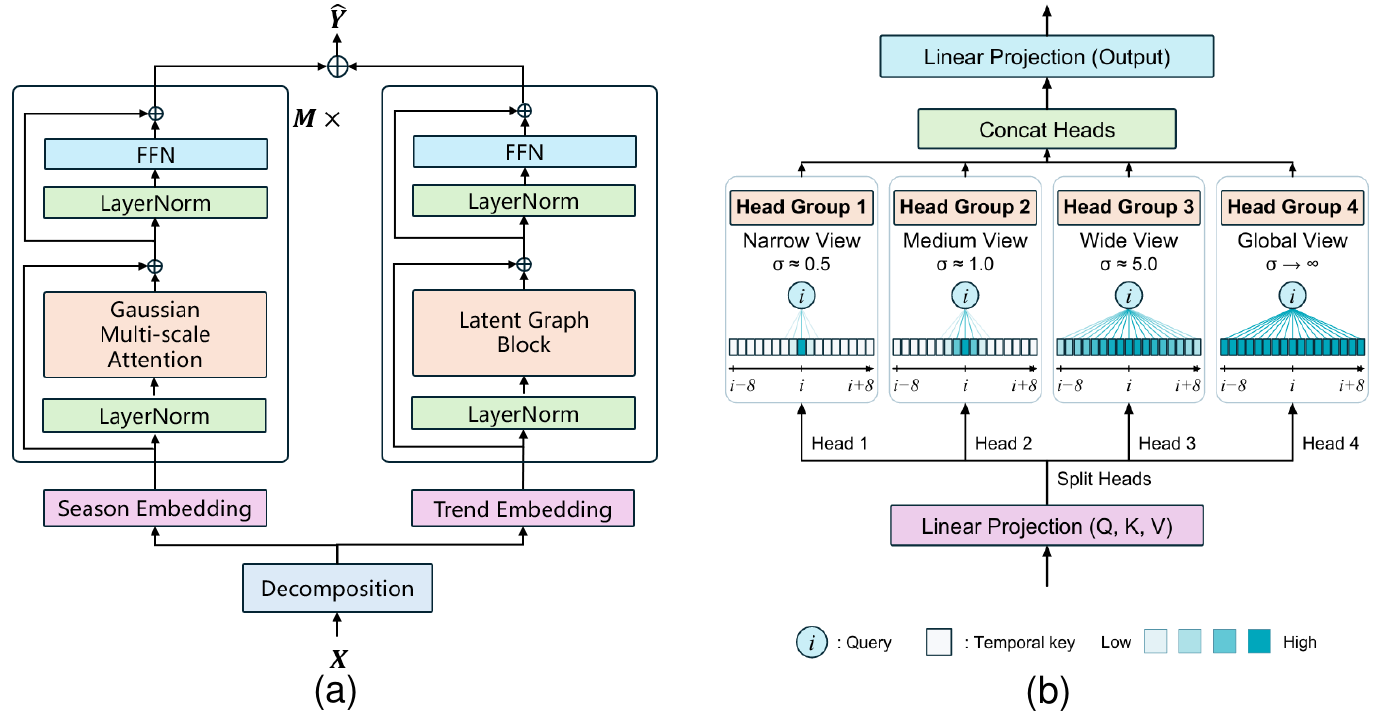}
    \caption{Overview of DST-Net. (a) Overall architecture. (b) Gaussian Multi-scale Attention (GMA).}
    \label{pipeline}
\end{figure}

\textbf{Overall Pipeline.} 
To effectively disentangle the conflicting inductive biases inherent in time series components, we propose the \textbf{D}ecomposed \textbf{S}patio-\textbf{T}emporal \textbf{Net}work (DST-Net). The overall pipeline is illustrated in Fig. \ref{pipeline}(a).

To decouple the complex temporal variations, we employ a standard series decomposition block. Following widely adopted practices~\cite{autoformer}, we utilize a moving average filter with kernel size $\kappa$ to extract the trend component. The seasonal component is then derived by subtracting the trend from the original series:
\begin{equation}
    \mathbf{X}_t = \text{AvgPool}(\mathbf{X}, \kappa), \quad \mathbf{X}_s = \mathbf{X} - \mathbf{X}_t,
\end{equation}
where $\mathbf{X}_t, \mathbf{X}_s \in \mathbb{R}^{L \times N}$ represent the trend and seasonal components, respectively.

Unlike previous methods that employ homogeneous networks for all components, DST-Net processes the decomposed series through two parallel branches tailored to their specific inductive biases. For the \textbf{Trend Branch (Spatial Domain)}, recognizing that the trend component represents the macroscopic system state characterized by temporal redundancy but strong spatial coupling, we introduce the Latent Graph Block. This branch compresses $\mathbf{X}_t$ into a low-dimensional latent space to eliminate temporal redundancy, explicitly models global topological dependencies to capture system-wide co-movements, and subsequently reconstructs the future trend $\hat{\mathbf{Y}}_t$. Simultaneously, the \textbf{Seasonal Branch (Temporal Domain)} addresses the seasonal component driven by local continuity. We design the Gaussian Multi-scale Attention to inject Gaussian priors, enforcing the model to focus on multi-granular local contexts. This mechanism effectively filters out global noise and distractions, leading to an accurate forecast of the future seasonality $\hat{\mathbf{Y}}_s$.

Finally, the predictions from both branches are aggregated to form the normalized forecast.
\begin{equation}
    \hat{\mathbf{Y}} = \hat{\mathbf{Y}}_t + \hat{\mathbf{Y}}_s.
\end{equation}

This disentangled pipeline ensures that each component is modeled with its optimal inductive bias, leading to superior representation capability and forecasting accuracy.

\subsection{Seasonal Branch: Gaussian Multi-scale Attention}
\label{sec:seasonal_branch}

\subsubsection{Motivation}

\begin{figure}[t]
    \begin{center}
    \includegraphics[width=\linewidth]{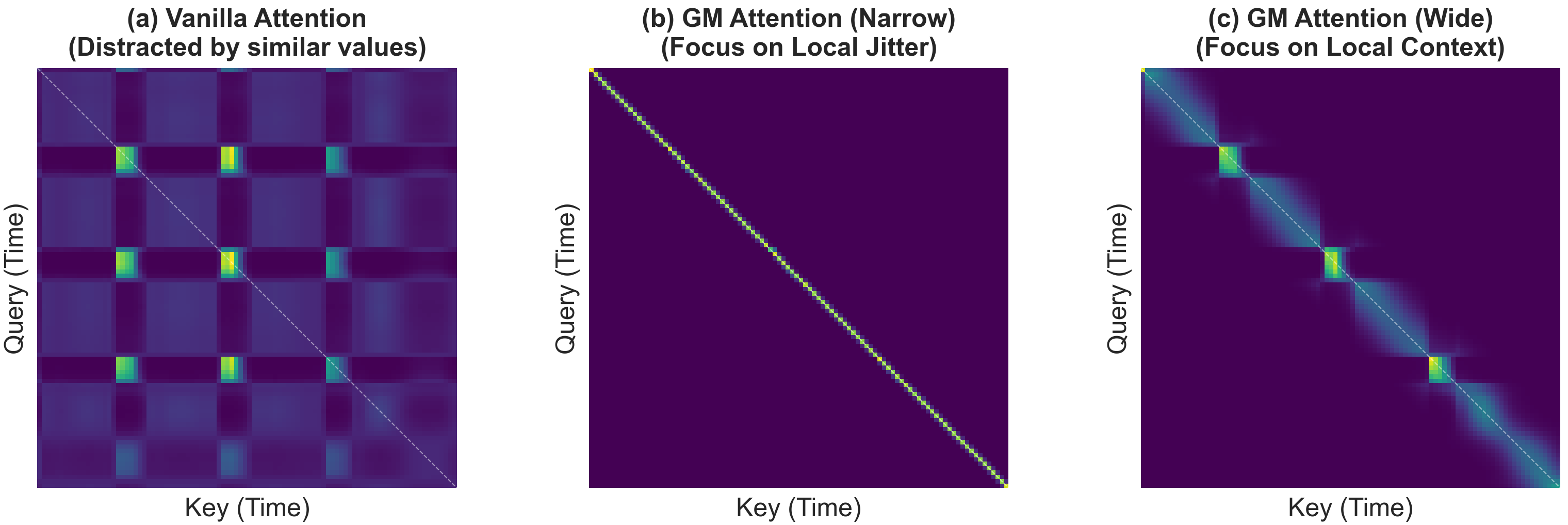}
    \caption{Visualization of Different Attention Maps.
    }\label{season_attn}
    \end{center}
\end{figure}

The seasonal component $\mathbf{X}_s$ is characterized by high-frequency fluctuations and recurring periodic patterns. Vanilla Transformers, which rely on a global receptive field and point-wise dot-product similarity, are limited for modeling such data. The limitation stems from the mismatch between \textit{numerical similarity} and \textit{state similarity}. In periodic sequences, temporally distant yet numerically similar points predominantly occupy distinct phases (e.g., rising vs. falling edges); however, vanilla attention assigns excessive weights to these points, resulting in the phenomenon of Attention Distraction. As visualized in Fig. \ref{season_attn}(a), this manifests as a scattered, checkerboard-like attention map. This global noise causes the model to over-smooth representations by aggregating irrelevant historical information, thereby losing the ability to distinguish subtle local variations and phase transitions. 

To resolve this ambiguity, we argue that \textit{temporal locality} is the critical inductive bias for seasonal modeling. The physical state of a seasonal point is defined not just by its value, but by its local continuity and trend, which can be inferred exclusively from its immediate context. As demonstrated in Fig. \ref{season_attn}(b) and (c), enforcing a local focus, either narrow for high-frequency jitter or wide for local context, effectively filters out these distant distractions and enables the model to accurately capture the continuous evolution of the waveform.

\subsubsection{Gaussian Multi-scale Attention}
Based on the insight that local context provides an important cue for distinguishing temporal phases, we propose the \textbf{Gaussian Multi-scale Attention (GMA)}. Instead of utilizing a hard-coded local window or allowing unconstrained global search, we inject learnable Gaussian priors into the attention mechanism to modulate the receptive field adaptively.

Formally, given the query $\mathbf{Q}$, key $\mathbf{K}$, and value $\mathbf{V}$ matrices, the standard attention score $S_{i,j}$ between query $i$ and key $j$ is computed as a scaled dot-product. In GMA, we introduce a distance-based Gaussian bias $G_{i,j}$ to the score:
\begin{equation}
    \text{Attention}(\mathbf{Q}, \mathbf{K}, \mathbf{V}) = \text{Softmax}\left(\frac{\mathbf{Q}\mathbf{K}^\top}{\sqrt{d_k}} + \mathbf{G}\right)\mathbf{V},
\end{equation}
where the Gaussian bias $\mathbf{G} \in \mathbb{R}^{L \times L}$ is defined based on the temporal distance $|i-j|$:
\begin{equation}
    G_{i,j} = -\frac{(i-j)^2}{2\sigma^2 + \epsilon}.
\end{equation}
Here, $\sigma$ is a learnable parameter that controls the bandwidth (i.e., the effective receptive field) of the attention head, and $\epsilon$ is a small constant for numerical stability. This additive bias enforces a soft locality constraint: when $\sigma$ is small, $G_{i,j}$ decays rapidly as the distance increases, forcing the attention to focus on immediate neighbors; when $\sigma$ is large, the mechanism approximates global attention.

To capture temporal patterns at various scales simultaneously, we adopt a multi-head splitting strategy as illustrated in Fig. \ref{pipeline}(b). Specifically, we partition the $H$ attention heads into four distinct groups, each initialized with a specific bandwidth $\sigma$ to target a unique temporal granularity. The \textbf{Narrow View} ($\sigma \approx 0.5$) restricts the receptive field to the immediate vicinity, focusing on high-frequency jitters and derivatives to determine instantaneous trends. The \textbf{Medium View} ($\sigma \approx 1.0$) captures local substructures and short-term continuity, while the \textbf{Wide View} ($\sigma \approx 5.0$) expands the focus to encompass meaningful local periodic contexts, such as significant portions of a cycle. Finally, to ensure the model retains the capacity for long-range dependency modeling, we allocate a \textbf{Global View} ($\sigma \to \infty$) that operates without Gaussian constraints. By explicitly assigning these distinct roles, this mechanism effectively suppresses distractions across multiple scales.

\subsection{Trend Branch: Latent Graph Block}
\label{sec:trend_branch}

Trend trajectories evolve smoothly and may co-vary across variates.
The \textbf{Latent Graph Block (LGB)} summarizes each variate's history
as a latent node and applies sparse cross-variate attention to capture
these dependencies.

Given $\mathbf{X}_t\in\mathbb{R}^{L\times N}$, the node representations
are obtained through a shared linear projection:
\begin{equation}
    \mathbf{Z}^{(0)}
    = \phi_{\mathrm{in}}(\mathbf{X}_t^\top)
    \in\mathbb{R}^{N\times d},
\end{equation}
where $\phi_{\mathrm{in}}:\mathbb{R}^{L}\rightarrow\mathbb{R}^{d}$.
For each head $h=1,\ldots,H_g$, learned linear projections map
$\operatorname{LN}(\mathbf{Z}^{(0)})$ to
$\mathbf{Q}^{(h)},\mathbf{K}^{(h)},\mathbf{V}^{(h)}
\in\mathbb{R}^{N\times d_h}$.
Neighboring features are aggregated using Top-$K$ attention:
\begin{equation}
    \mathbf{M}^{(h)}
    =
    \operatorname{Softmax}
    \left(
        \mathcal{M}_{K}
        \left(
            \frac{
                \mathbf{Q}^{(h)}
                (\mathbf{K}^{(h)})^\top
            }{\sqrt{d_h}}
        \right)
    \right)
    \mathbf{V}^{(h)},
\end{equation}
where $\mathcal{M}_{K}$ retains the $K$ largest scores in each row
and sets the remaining entries to $-\infty$ before row-wise softmax.

The head outputs are concatenated and projected back to $d$ dimensions
through a residual update:
\begin{equation}
    \mathbf{Z}_{a}
    =
    \mathbf{Z}^{(0)}
    +
    \phi_{\mathrm{out}}
    \left(
        \operatorname{Concat}
        [\mathbf{M}^{(1)},\ldots,\mathbf{M}^{(H_g)}]
    \right).
\end{equation}
A feed-forward block with residual connections and layer normalization
produces $\mathbf{Z}_t\in\mathbb{R}^{N\times d}$.
A shared prediction head
$\phi_{\mathrm{pred}}:\mathbb{R}^{d}\rightarrow\mathbb{R}^{T}$
then yields
$\hat{\mathbf{Y}}_t=\phi_{\mathrm{pred}}(\mathbf{Z}_t)^\top
\in\mathbb{R}^{T\times N}$.

\section{Experiments}

\subsection{Experimental Setup}

We evaluate DST-Net on both long-term and short-term forecasting tasks.
Long-term experiments use seven benchmarks: ETTh1, ETTh2, ETTm1, ETTm2,
Exchange, Traffic, and Weather, with prediction horizons of 96, 192, 336,
and 720. Short-term experiments use PEMS04 with an input length of 96
and a prediction horizon of 12. Data preprocessing and training,
validation, and test splits follow previous studies
\cite{patchtst,itransformer}.

We compare DST-Net with representative models, including TimeMixer
\cite{timemixer}, iTransformer \cite{itransformer}, and PatchTST
\cite{patchtst}, covering multi-scale, variate-wise, and patch-based
modeling. DST-Net is implemented in PyTorch, and all experiments run on
a single NVIDIA H20 GPU with 144\,GB of memory. All methods use identical data splits and preprocessing. Unless otherwise
specified, the unified configurations follow TimeMixer \cite{timemixer}.

\subsection{Evaluation Metrics}

For long-term time series forecasting, we employ Mean Squared Error (MSE) and Mean Absolute Error (MAE) as the evaluation metrics. For short-term forecasting, following the standard settings of previous studies, we adopt MAE, Root Mean Squared Error (RMSE), and Mean Absolute Percentage Error (MAPE). Lower values indicate better forecasting performance for all evaluation metrics.

\subsection{Long-term Forecasting}
Table~\ref{long_result} shows that DST-Net achieves leading average
performance on most benchmarks and consistently ranks among the strongest
methods across prediction horizons. Its competitive MSE and MAE at both
shorter and longer horizons indicate that the gains persist across the
evaluated forecasting ranges.

\begin{table}[h]
  \caption{Long-term forecasting results. We average the results across 4 prediction lengths: $\{96, 192, 336, 720\}$. The best performance is highlighted in \textcolor{red}{\textbf{red}}, and the second-best is \textcolor{blue}{\underline{underlined}}.}
  \label{long_result}
  \centering
  \resizebox{1.0\columnwidth}{!}{
  \begin{threeparttable}
  \begin{small}
  \renewcommand{\multirowsetup}{\centering}
  \setlength{\tabcolsep}{1pt}
  \begin{tabular}{c|c|cc|cc|cc|cc|cc|cc|cc|cc|cc|cc|cc|cc}
    \toprule
    \multicolumn{2}{c}{\multirow{2}{*}{Models}} & \multicolumn{2}{c}{\rotatebox{0}{\scalebox{0.8}{\textbf{DST-Net}}}} &
    \multicolumn{2}{c}{\rotatebox{0}{\scalebox{0.8}{\textbf{TimeMixer}}}} &
    \multicolumn{2}{c}{\rotatebox{0}{\scalebox{0.8}{iTransformer}}} &
    \multicolumn{2}{c}{\rotatebox{0}{\scalebox{0.8}{PatchTST}}} &
    \multicolumn{2}{c}{\rotatebox{0}{\scalebox{0.8}{Crossformer}}} &
    \multicolumn{2}{c}{\rotatebox{0}{\scalebox{0.8}{TiDE}}} &
    \multicolumn{2}{c}{\rotatebox{0}{\scalebox{0.8}{TimesNet}}} & 
    \multicolumn{2}{c}{\rotatebox{0}{\scalebox{0.8}{DLinear}}} &
    \multicolumn{2}{c}{\rotatebox{0}{\scalebox{0.8}{SCINet}}}&
    \multicolumn{2}{c}{\rotatebox{0}{\scalebox{0.8}{FEDformer}}} & 
    \multicolumn{2}{c}{\rotatebox{0}{\scalebox{0.8}{Stationary}}} & 
    \multicolumn{2}{c}{\rotatebox{0}{\scalebox{0.8}{Autoformer}}}  \\
    \multicolumn{2}{c}{ }&
    \multicolumn{2}{c}{\scalebox{0.8}{\textbf{(Ours)}}} &
    \multicolumn{2}{c}{\scalebox{0.8}{\cite{timemixer}}} &
    \multicolumn{2}{c}{\scalebox{0.8}{\cite{itransformer}}} & 
    \multicolumn{2}{c}{\scalebox{0.8}{\cite{patchtst}}} & 
    \multicolumn{2}{c}{\scalebox{0.8}{\cite{crossformer}}} & 
    \multicolumn{2}{c}{\scalebox{0.8}{\cite{tilde}}} & 
    \multicolumn{2}{c}{\scalebox{0.8}{\cite{timesnet}}} & 
    \multicolumn{2}{c}{\scalebox{0.8}{\cite{dlinear}}} &
    \multicolumn{2}{c}{\scalebox{0.8}{\cite{scinet}}} & 
    \multicolumn{2}{c}{\scalebox{0.8}{\cite{fedformer}}} &
    \multicolumn{2}{c}{\scalebox{0.8}{\cite{NonstationaryTR}}} &
    \multicolumn{2}{c}{\scalebox{0.8}{\cite{autoformer}}} \\
    \midrule

    \multirow{5}{*}{\rotatebox{90}{\scalebox{0.95}{Exchange}}} 
    & \scalebox{0.85}{96}
    & \boldres{\scalebox{0.85}{0.082}}
    & \boldres{\scalebox{0.85}{0.200}}
    & \scalebox{0.85}{0.090}
    & \scalebox{0.85}{0.235}
    & \secondres{\scalebox{0.85}{0.086}}
    & \scalebox{0.85}{0.206}
    & \scalebox{0.85}{0.088}
    & \secondres{\scalebox{0.85}{0.205}}
    & \scalebox{0.85}{0.256}
    & \scalebox{0.85}{0.367}
    & \scalebox{0.85}{0.094}
    & \scalebox{0.85}{0.218}
    & \scalebox{0.85}{0.107}
    & \scalebox{0.85}{0.234}
    & \scalebox{0.85}{0.088}
    & \scalebox{0.85}{0.218}
    & \scalebox{0.85}{0.267}
    & \scalebox{0.85}{0.396}
    & \scalebox{0.85}{0.148}
    & \scalebox{0.85}{0.278}
    & \scalebox{0.85}{0.111}
    & \scalebox{0.85}{0.237}
    & \scalebox{0.85}{0.197}
    & \scalebox{0.85}{0.323}
    \\

    & \scalebox{0.85}{192}
    & \boldres{\scalebox{0.85}{0.168}}
    & \boldres{\scalebox{0.85}{0.294}}
    & \scalebox{0.85}{0.187}
    & \scalebox{0.85}{0.343}
    & \scalebox{0.85}{0.177}
    & \secondres{\scalebox{0.85}{0.299}}
    & \secondres{\scalebox{0.85}{0.176}}
    & \secondres{\scalebox{0.85}{0.299}}
    & \scalebox{0.85}{0.470}
    & \scalebox{0.85}{0.509}
    & \scalebox{0.85}{0.184}
    & \scalebox{0.85}{0.307}
    & \scalebox{0.85}{0.226}
    & \scalebox{0.85}{0.344}
    & \secondres{\scalebox{0.85}{0.176}}
    & \scalebox{0.85}{0.315}
    & \scalebox{0.85}{0.351}
    & \scalebox{0.85}{0.459}
    & \scalebox{0.85}{0.271}
    & \scalebox{0.85}{0.315}
    & \scalebox{0.85}{0.219}
    & \scalebox{0.85}{0.335}
    & \scalebox{0.85}{0.300}
    & \scalebox{0.85}{0.369}
    \\

    & \scalebox{0.85}{336}
    & \scalebox{0.85}{0.320}
    & \secondres{\scalebox{0.85}{0.409}}
    & \scalebox{0.85}{0.353}
    & \scalebox{0.85}{0.473}
    & \scalebox{0.85}{0.331}
    & \scalebox{0.85}{0.417}
    & \boldres{\scalebox{0.85}{0.301}}
    & \boldres{\scalebox{0.85}{0.397}}
    & \scalebox{0.85}{1.268}
    & \scalebox{0.85}{0.883}
    & \scalebox{0.85}{0.349}
    & \scalebox{0.85}{0.431}
    & \scalebox{0.85}{0.367}
    & \scalebox{0.85}{0.448}
    & \secondres{\scalebox{0.85}{0.313}}
    & \scalebox{0.85}{0.427}
    & \scalebox{0.85}{1.324}
    & \scalebox{0.85}{0.853}
    & \scalebox{0.85}{0.460}
    & \scalebox{0.85}{0.427}
    & \scalebox{0.85}{0.421}
    & \scalebox{0.85}{0.476}
    & \scalebox{0.85}{0.509}
    & \scalebox{0.85}{0.524}
    \\

    & \scalebox{0.85}{720}
    & \boldres{\scalebox{0.85}{0.785}}
    & \boldres{\scalebox{0.85}{0.669}}
    & \scalebox{0.85}{0.934}
    & \scalebox{0.85}{0.761}
    & \scalebox{0.85}{0.847}
    & \secondres{\scalebox{0.85}{0.691}}
    & \scalebox{0.85}{0.901}
    & \scalebox{0.85}{0.714}
    & \scalebox{0.85}{1.767}
    & \scalebox{0.85}{1.068}
    & \scalebox{0.85}{0.852}
    & \scalebox{0.85}{0.698}
    & \scalebox{0.85}{0.964}
    & \scalebox{0.85}{0.746}
    & \secondres{\scalebox{0.85}{0.839}}
    & \scalebox{0.85}{0.695}
    & \scalebox{0.85}{1.058}
    & \scalebox{0.85}{0.797}
    & \scalebox{0.85}{1.195}
    & \scalebox{0.85}{0.695}
    & \scalebox{0.85}{1.092}
    & \scalebox{0.85}{0.769}
    & \scalebox{0.85}{1.447}
    & \scalebox{0.85}{0.941}
    \\

    \cmidrule(lr){2-26}

    & \scalebox{0.85}{Avg}
    & \boldres{\scalebox{0.8}{0.339}}
    & \boldres{\scalebox{0.8}{0.393}}
    & \scalebox{0.85}{0.391}
    & \scalebox{0.85}{0.453}
    & \scalebox{0.85}{0.360}
    & \secondres{\scalebox{0.85}{0.403}}
    & \scalebox{0.85}{0.367}
    & \scalebox{0.85}{0.404}
    & \scalebox{0.85}{0.940}
    & \scalebox{0.85}{0.707}
    & \scalebox{0.85}{0.370}
    & \scalebox{0.85}{0.413}
    & \scalebox{0.85}{0.416}
    & \scalebox{0.85}{0.443}
    & \secondres{\scalebox{0.85}{0.354}}
    & \scalebox{0.85}{0.414}
    & \scalebox{0.85}{0.750}
    & \scalebox{0.85}{0.626}
    & \scalebox{0.85}{0.519}
    & \scalebox{0.85}{0.429}
    & \scalebox{0.85}{0.461}
    & \scalebox{0.85}{0.454}
    & \scalebox{0.85}{0.613}
    & \scalebox{0.85}{0.539}
    \\
\midrule

    \multirow{5}{*}{\rotatebox{90}{\scalebox{0.95}{ETTh1}}} 
    & \scalebox{0.85}{96}
    & \boldres{\scalebox{0.85}{0.375}} & \boldres{\scalebox{0.85}{0.399}}
    & \boldres{\scalebox{0.85}{0.375}} & \secondres{\scalebox{0.85}{0.400}}
    & \scalebox{0.85}{0.386} & \scalebox{0.85}{0.405}
    & \scalebox{0.85}{0.460} & \scalebox{0.85}{0.447}
    & \scalebox{0.85}{0.423} & \scalebox{0.85}{0.448}
    & \scalebox{0.85}{0.479} & \scalebox{0.85}{0.464}
    & \secondres{\scalebox{0.85}{0.384}} & \scalebox{0.85}{0.402}
    & \scalebox{0.85}{0.397} & \scalebox{0.85}{0.412}
    & \scalebox{0.85}{0.654} & \scalebox{0.85}{0.599}
    & \scalebox{0.85}{0.395} & \scalebox{0.85}{0.424}
    & \scalebox{0.85}{0.513} & \scalebox{0.85}{0.491}
    & \scalebox{0.85}{0.449} & \scalebox{0.85}{0.459}
    \\
    & \scalebox{0.85}{192}
    & \boldres{\scalebox{0.85}{0.415}} & \secondres{\scalebox{0.85}{0.428}}
    & \secondres{\scalebox{0.85}{0.429}} & \boldres{\scalebox{0.85}{0.421}}
    & \scalebox{0.85}{0.441} & \scalebox{0.85}{0.436}
    & \scalebox{0.85}{0.512} & \scalebox{0.85}{0.477}
    & \scalebox{0.85}{0.471} & \scalebox{0.85}{0.474}
    & \scalebox{0.85}{0.525} & \scalebox{0.85}{0.492}
    & \scalebox{0.85}{0.436} & \scalebox{0.85}{0.429}
    & \scalebox{0.85}{0.446} & \scalebox{0.85}{0.441}
    & \scalebox{0.85}{0.719} & \scalebox{0.85}{0.631}
    & \scalebox{0.85}{0.469} & \scalebox{0.85}{0.470}
    & \scalebox{0.85}{0.534} & \scalebox{0.85}{0.504}
    & \scalebox{0.85}{0.500} & \scalebox{0.85}{0.482}
    \\
    & \scalebox{0.85}{336}
    & \boldres{\scalebox{0.85}{0.439}} & \boldres{\scalebox{0.85}{0.437}}
    & \secondres{\scalebox{0.85}{0.484}} & \secondres{\scalebox{0.85}{0.458}}
    & \scalebox{0.85}{0.487} & \secondres{\scalebox{0.85}{0.458}}
    & \scalebox{0.85}{0.546} & \scalebox{0.85}{0.496}
    & \scalebox{0.85}{0.570} & \scalebox{0.85}{0.546}
    & \scalebox{0.85}{0.565} & \scalebox{0.85}{0.515}
    & \scalebox{0.85}{0.491} & \scalebox{0.85}{0.469}
    & \scalebox{0.85}{0.489} & \scalebox{0.85}{0.467}
    & \scalebox{0.85}{0.778} & \scalebox{0.85}{0.659}
    & \scalebox{0.85}{0.530} & \scalebox{0.85}{0.499}
    & \scalebox{0.85}{0.588} & \scalebox{0.85}{0.535}
    & \scalebox{0.85}{0.521} & \scalebox{0.85}{0.496}
    \\
    & \scalebox{0.85}{720}
    & \boldres{\scalebox{0.85}{0.444}} & \boldres{\scalebox{0.85}{0.457}}
    & \secondres{\scalebox{0.85}{0.498}} & \secondres{\scalebox{0.85}{0.482}}
    & \scalebox{0.85}{0.503} & \scalebox{0.85}{0.491}
    & \scalebox{0.85}{0.544} & \scalebox{0.85}{0.517}
    & \scalebox{0.85}{0.653} & \scalebox{0.85}{0.621}
    & \scalebox{0.85}{0.594} & \scalebox{0.85}{0.558}
    & \scalebox{0.85}{0.521} & \scalebox{0.85}{0.500}
    & \scalebox{0.85}{0.513} & \scalebox{0.85}{0.510}
    & \scalebox{0.85}{0.836} & \scalebox{0.85}{0.699}
    & \scalebox{0.85}{0.598} & \scalebox{0.85}{0.544}
    & \scalebox{0.85}{0.643} & \scalebox{0.85}{0.616}
    & \scalebox{0.85}{0.514} & \scalebox{0.85}{0.512}
    \\
    \cmidrule(lr){2-26}
    & \scalebox{0.85}{Avg}
    & \boldres{\scalebox{0.85}{0.418}} & \boldres{\scalebox{0.85}{0.430}}
    & \secondres{\scalebox{0.85}{0.447}} & \secondres{\scalebox{0.85}{0.440}}
    & \scalebox{0.85}{0.454} & \scalebox{0.85}{0.447}
    & \scalebox{0.85}{0.516} & \scalebox{0.85}{0.484}
    & \scalebox{0.85}{0.529} & \scalebox{0.85}{0.522}
    & \scalebox{0.85}{0.541} & \scalebox{0.85}{0.507}
    & \scalebox{0.85}{0.458} & \scalebox{0.85}{0.450}
    & \scalebox{0.85}{0.461} & \scalebox{0.85}{0.457}
    & \scalebox{0.85}{0.747} & \scalebox{0.85}{0.647}
    & \scalebox{0.85}{0.498} & \scalebox{0.85}{0.484}
    & \scalebox{0.85}{0.570} & \scalebox{0.85}{0.537}
    & \scalebox{0.85}{0.496} & \scalebox{0.85}{0.487}
    \\
\midrule

\multirow{5}{*}{\rotatebox{90}{\scalebox{0.95}{ETTh2}}} 
    & \scalebox{0.85}{96}
    & \secondres{\scalebox{0.85}{0.296}} & \secondres{\scalebox{0.85}{0.345}}
    & \boldres{\scalebox{0.85}{0.289}} & \boldres{\scalebox{0.85}{0.341}}
    & \scalebox{0.85}{0.297} & \scalebox{0.85}{0.349}
    & \scalebox{0.85}{0.308} & \scalebox{0.85}{0.355}
    & \scalebox{0.85}{0.745} & \scalebox{0.85}{0.584}
    & \scalebox{0.85}{0.400} & \scalebox{0.85}{0.440}
    & \scalebox{0.85}{0.340} & \scalebox{0.85}{0.374}
    & \scalebox{0.85}{0.340} & \scalebox{0.85}{0.394}
    & \scalebox{0.85}{0.707} & \scalebox{0.85}{0.621}
    & \scalebox{0.85}{0.358} & \scalebox{0.85}{0.397}
    & \scalebox{0.85}{0.476} & \scalebox{0.85}{0.458}
    & \scalebox{0.85}{0.346} & \scalebox{0.85}{0.388}
    \\
    & \scalebox{0.85}{192}
    & \boldres{\scalebox{0.85}{0.367}} & \boldres{\scalebox{0.85}{0.390}}
    & \secondres{\scalebox{0.85}{0.372}} & \secondres{\scalebox{0.85}{0.392}}
    & \scalebox{0.85}{0.380} & \scalebox{0.85}{0.400}
    & \scalebox{0.85}{0.393} & \scalebox{0.85}{0.405}
    & \scalebox{0.85}{0.877} & \scalebox{0.85}{0.656}
    & \scalebox{0.85}{0.528} & \scalebox{0.85}{0.509}
    & \scalebox{0.85}{0.402} & \scalebox{0.85}{0.414}
    & \scalebox{0.85}{0.482} & \scalebox{0.85}{0.479}
    & \scalebox{0.85}{0.860} & \scalebox{0.85}{0.689}
    & \scalebox{0.85}{0.429} & \scalebox{0.85}{0.439}
    & \scalebox{0.85}{0.512} & \scalebox{0.85}{0.493}
    & \scalebox{0.85}{0.456} & \scalebox{0.85}{0.452}
    \\
    & \scalebox{0.85}{336}
    & \boldres{\scalebox{0.85}{0.370}} & \boldres{\scalebox{0.85}{0.405}}
    & \secondres{\scalebox{0.85}{0.386}} & \secondres{\scalebox{0.85}{0.414}}
    & \scalebox{0.85}{0.428} & \scalebox{0.85}{0.432}
    & \scalebox{0.85}{0.427} & \scalebox{0.85}{0.436}
    & \scalebox{0.85}{1.043} & \scalebox{0.85}{0.731}
    & \scalebox{0.85}{0.643} & \scalebox{0.85}{0.571}
    & \scalebox{0.85}{0.452} & \scalebox{0.85}{0.452}
    & \scalebox{0.85}{0.591} & \scalebox{0.85}{0.541}
    & \scalebox{0.85}{1.000} & \scalebox{0.85}{0.744}
    & \scalebox{0.85}{0.496} & \scalebox{0.85}{0.487}
    & \scalebox{0.85}{0.552} & \scalebox{0.85}{0.551}
    & \scalebox{0.85}{0.482} & \scalebox{0.85}{0.486}
    \\
    & \scalebox{0.85}{720}
    & \boldres{\scalebox{0.85}{0.407}} & \boldres{\scalebox{0.85}{0.434}}
    & \secondres{\scalebox{0.85}{0.412}} & \boldres{\scalebox{0.85}{0.434}}
    & \scalebox{0.85}{0.427} & \secondres{\scalebox{0.85}{0.445}}
    & \scalebox{0.85}{0.436} & \scalebox{0.85}{0.450}
    & \scalebox{0.85}{1.104} & \scalebox{0.85}{0.763}
    & \scalebox{0.85}{0.874} & \scalebox{0.85}{0.679}
    & \scalebox{0.85}{0.462} & \scalebox{0.85}{0.468}
    & \scalebox{0.85}{0.839} & \scalebox{0.85}{0.661}
    & \scalebox{0.85}{1.249} & \scalebox{0.85}{0.838}
    & \scalebox{0.85}{0.463} & \scalebox{0.85}{0.474}
    & \scalebox{0.85}{0.562} & \scalebox{0.85}{0.560}
    & \scalebox{0.85}{0.515} & \scalebox{0.85}{0.511}
    \\
    \cmidrule(lr){2-26}
    & \scalebox{0.85}{Avg}
    & \boldres{\scalebox{0.85}{0.360}} & \boldres{\scalebox{0.85}{0.394}}
    & \secondres{\scalebox{0.85}{0.364}} & \secondres{\scalebox{0.85}{0.395}}
    & \scalebox{0.85}{0.383} & \scalebox{0.85}{0.407}
    & \scalebox{0.85}{0.391} & \scalebox{0.85}{0.411}
    & \scalebox{0.85}{0.942} & \scalebox{0.85}{0.684}
    & \scalebox{0.85}{0.611} & \scalebox{0.85}{0.550}
    & \scalebox{0.85}{0.414} & \scalebox{0.85}{0.427}
    & \scalebox{0.85}{0.563} & \scalebox{0.85}{0.519}
    & \scalebox{0.85}{0.954} & \scalebox{0.85}{0.723}
    & \scalebox{0.85}{0.437} & \scalebox{0.85}{0.449}
    & \scalebox{0.85}{0.526} & \scalebox{0.85}{0.516}
    & \scalebox{0.85}{0.450} & \scalebox{0.85}{0.459}
    \\
\midrule

\multirow{5}{*}{\rotatebox{90}{\scalebox{0.95}{ETTm1}}} 
    & \scalebox{0.85}{96}
    & \boldres{\scalebox{0.85}{0.319}} & \secondres{\scalebox{0.85}{0.358}}
    & \secondres{\scalebox{0.85}{0.320}} & \boldres{\scalebox{0.85}{0.357}}
    & \scalebox{0.85}{0.334} & \scalebox{0.85}{0.368}
    & \scalebox{0.85}{0.352} & \scalebox{0.85}{0.374}
    & \scalebox{0.85}{0.404} & \scalebox{0.85}{0.426}
    & \scalebox{0.85}{0.364} & \scalebox{0.85}{0.387}
    & \scalebox{0.85}{0.338} & \scalebox{0.85}{0.375}
    & \scalebox{0.85}{0.346} & \scalebox{0.85}{0.374}
    & \scalebox{0.85}{0.418} & \scalebox{0.85}{0.438}
    & \scalebox{0.85}{0.379} & \scalebox{0.85}{0.419}
    & \scalebox{0.85}{0.386} & \scalebox{0.85}{0.398}
    & \scalebox{0.85}{0.505} & \scalebox{0.85}{0.475}
    \\
    & \scalebox{0.85}{192}
    & \boldres{\scalebox{0.85}{0.359}} & \boldres{\scalebox{0.85}{0.381}}
    & \secondres{\scalebox{0.85}{0.361}} & \boldres{\scalebox{0.85}{0.381}}
    & \scalebox{0.85}{0.377} & \scalebox{0.85}{0.391}
    & \scalebox{0.85}{0.390} & {\scalebox{0.85}{0.393}}
    & \scalebox{0.85}{0.450} & \scalebox{0.85}{0.451}
    & \scalebox{0.85}{0.398} & \scalebox{0.85}{0.404}
    & \scalebox{0.85}{0.374} & \secondres{\scalebox{0.85}{0.387}}
    & \scalebox{0.85}{0.382} & \scalebox{0.85}{0.391}
    & \scalebox{0.85}{0.439} & \scalebox{0.85}{0.450}
    & \scalebox{0.85}{0.426} & \scalebox{0.85}{0.441}
    & \scalebox{0.85}{0.459} & \scalebox{0.85}{0.444}
    & \scalebox{0.85}{0.553} & \scalebox{0.85}{0.496}
    \\
    & \scalebox{0.85}{336}
    & \boldres{\scalebox{0.85}{0.389}} & \secondres{\scalebox{0.85}{0.405}}
    & \secondres{\scalebox{0.85}{0.390}} & \boldres{\scalebox{0.85}{0.404}}
    & \scalebox{0.85}{0.426} & \scalebox{0.85}{0.420}
    & \scalebox{0.85}{0.421} & \scalebox{0.85}{0.414}
    & \scalebox{0.85}{0.532} & \scalebox{0.85}{0.515}
    & \scalebox{0.85}{0.428} & \scalebox{0.85}{0.425}
    & \scalebox{0.85}{0.410} & \scalebox{0.85}{0.411}
    & \scalebox{0.85}{0.415} & \scalebox{0.85}{0.415}
    & \scalebox{0.85}{0.490} & \scalebox{0.85}{0.485}
    & \scalebox{0.85}{0.445} & \scalebox{0.85}{0.459}
    & \scalebox{0.85}{0.495} & \scalebox{0.85}{0.464}
    & \scalebox{0.85}{0.621} & \scalebox{0.85}{0.537}
    \\
    & \scalebox{0.85}{720}
    & \boldres{\scalebox{0.85}{0.451}} & \secondres{\scalebox{0.85}{0.445}}
    & \secondres{\scalebox{0.85}{0.454}} & \boldres{\scalebox{0.85}{0.441}}
    & \scalebox{0.85}{0.491} & \scalebox{0.85}{0.459}
    & \scalebox{0.85}{0.462} & \scalebox{0.85}{0.449}
    & \scalebox{0.85}{0.666} & \scalebox{0.85}{0.589}
    & \scalebox{0.85}{0.487} & \scalebox{0.85}{0.461}
    & \scalebox{0.85}{0.478} & \scalebox{0.85}{0.450}
    & \scalebox{0.85}{0.473} & \scalebox{0.85}{0.451}
    & \scalebox{0.85}{0.595} & \scalebox{0.85}{0.550}
    & \scalebox{0.85}{0.543} & \scalebox{0.85}{0.490}
    & \scalebox{0.85}{0.585} & \scalebox{0.85}{0.516}
    & \scalebox{0.85}{0.671} & \scalebox{0.85}{0.561}
    \\
    \cmidrule(lr){2-26}
    & \scalebox{0.85}{Avg}
    & \boldres{\scalebox{0.85}{0.379}} & \secondres{\scalebox{0.85}{0.397}}
    & \secondres{\scalebox{0.85}{0.381}} & \boldres{\scalebox{0.85}{0.395}}
    & \scalebox{0.85}{0.407} & \scalebox{0.85}{0.410}
    & \scalebox{0.85}{0.406} & \scalebox{0.85}{0.407}
    & \scalebox{0.85}{0.513} & \scalebox{0.85}{0.495}
    & \scalebox{0.85}{0.419} & \scalebox{0.85}{0.419}
    & \scalebox{0.85}{0.400} & \scalebox{0.85}{0.406}
    & \scalebox{0.85}{0.404} & \scalebox{0.85}{0.408}
    & \scalebox{0.85}{0.485} & \scalebox{0.85}{0.481}
    & \scalebox{0.85}{0.448} & \scalebox{0.85}{0.452}
    & \scalebox{0.85}{0.481} & \scalebox{0.85}{0.456}
    & \scalebox{0.85}{0.588} & \scalebox{0.85}{0.517}
    \\
\midrule

\multirow{5}{*}{\rotatebox{90}{\scalebox{0.95}{ETTm2}}} 
    & \scalebox{0.85}{96}
    & \secondres{\scalebox{0.85}{0.177}} & \secondres{\scalebox{0.85}{0.261}}
    & \boldres{\scalebox{0.85}{0.175}} & \boldres{\scalebox{0.85}{0.258}}
    & \scalebox{0.85}{0.180} & \scalebox{0.85}{0.264}
    & \scalebox{0.85}{0.183} & \scalebox{0.85}{0.270}
    & \scalebox{0.85}{0.287} & \scalebox{0.85}{0.366}
    & \scalebox{0.85}{0.207} & \scalebox{0.85}{0.305}
    & \scalebox{0.85}{0.187} & \scalebox{0.85}{0.267}
    & \scalebox{0.85}{0.193} & \scalebox{0.85}{0.293}
    & \scalebox{0.85}{0.286} & \scalebox{0.85}{0.377}
    & \scalebox{0.85}{0.203} & \scalebox{0.85}{0.287}
    & \scalebox{0.85}{0.192} & \scalebox{0.85}{0.274}
    & \scalebox{0.85}{0.255} & \scalebox{0.85}{0.339}
    \\
    & \scalebox{0.85}{192}
    & \secondres{\scalebox{0.85}{0.243}} & \secondres{\scalebox{0.85}{0.306}}
    & \boldres{\scalebox{0.85}{0.237}} & \boldres{\scalebox{0.85}{0.299}}
    & \scalebox{0.85}{0.250} & \scalebox{0.85}{0.309}
    & \scalebox{0.85}{0.255} & \scalebox{0.85}{0.314}
    & \scalebox{0.85}{0.414} & \scalebox{0.85}{0.492}
    & \scalebox{0.85}{0.290} & \scalebox{0.85}{0.364}
    & \scalebox{0.85}{0.249} & \scalebox{0.85}{0.309}
    & \scalebox{0.85}{0.284} & \scalebox{0.85}{0.361}
    & \scalebox{0.85}{0.399} & \scalebox{0.85}{0.445}
    & \scalebox{0.85}{0.269} & \scalebox{0.85}{0.328}
    & \scalebox{0.85}{0.280} & \scalebox{0.85}{0.339}
    & \scalebox{0.85}{0.281} & \scalebox{0.85}{0.340}
    \\
    & \scalebox{0.85}{336}
    & \secondres{\scalebox{0.85}{0.306}} & \secondres{\scalebox{0.85}{0.344}}
    & \boldres{\scalebox{0.85}{0.298}} & \boldres{\scalebox{0.85}{0.340}}
    & \scalebox{0.85}{0.311} & \scalebox{0.85}{0.348}
    & \scalebox{0.85}{0.309} & \scalebox{0.85}{0.347}
    & \scalebox{0.85}{0.597} & \scalebox{0.85}{0.542}
    & \scalebox{0.85}{0.377} & \scalebox{0.85}{0.422}
    & \scalebox{0.85}{0.321} & \scalebox{0.85}{0.351}
    & \scalebox{0.85}{0.382} & \scalebox{0.85}{0.429}
    & \scalebox{0.85}{0.637} & \scalebox{0.85}{0.591}
    & \scalebox{0.85}{0.325} & \scalebox{0.85}{0.366}
    & \scalebox{0.85}{0.334} & \scalebox{0.85}{0.361}
    & \scalebox{0.85}{0.339} & \scalebox{0.85}{0.372}
    \\
    & \scalebox{0.85}{720}
    & \secondres{\scalebox{0.85}{0.404}} & \secondres{\scalebox{0.85}{0.403}}
    & \boldres{\scalebox{0.85}{0.391}} & \boldres{\scalebox{0.85}{0.396}}
    & \scalebox{0.85}{0.412} & \scalebox{0.85}{0.407}
    & \scalebox{0.85}{0.412} & \scalebox{0.85}{0.404}
    & \scalebox{0.85}{1.730} & \scalebox{0.85}{1.042}
    & \scalebox{0.85}{0.558} & \scalebox{0.85}{0.524}
    & \scalebox{0.85}{0.408} & \secondres{\scalebox{0.85}{0.403}}
    & \scalebox{0.85}{0.558} & \scalebox{0.85}{0.525}
    & \scalebox{0.85}{0.960} & \scalebox{0.85}{0.735}
    & \scalebox{0.85}{0.421} & \scalebox{0.85}{0.415}
    & \scalebox{0.85}{0.417} & \scalebox{0.85}{0.413}
    & \scalebox{0.85}{0.433} & \scalebox{0.85}{0.432}
    \\
    \cmidrule(lr){2-26}
    & \scalebox{0.85}{Avg}
    & \secondres{\scalebox{0.85}{0.282}} & \secondres{\scalebox{0.85}{0.329}}
    & \boldres{\scalebox{0.85}{0.275}} & \boldres{\scalebox{0.85}{0.323}}
    & \scalebox{0.85}{0.288} & \scalebox{0.85}{0.332}
    & \scalebox{0.85}{0.290} & \scalebox{0.85}{0.334}
    & \scalebox{0.85}{0.757} & \scalebox{0.85}{0.610}
    & \scalebox{0.85}{0.358} & \scalebox{0.85}{0.404}
    & \scalebox{0.85}{0.291} & \scalebox{0.85}{0.333}
    & \scalebox{0.85}{0.354} & \scalebox{0.85}{0.402}
    & \scalebox{0.85}{0.571} & \scalebox{0.85}{0.537}
    & \scalebox{0.85}{0.305} & \scalebox{0.85}{0.349}
    & \scalebox{0.85}{0.306} & \scalebox{0.85}{0.347}
    & \scalebox{0.85}{0.327} & \scalebox{0.85}{0.371}
    \\
    \midrule
\multirow{5}{*}{\rotatebox{90}{\scalebox{0.95}{Weather}}} 
    & \scalebox{0.85}{96}
    & \boldres{\scalebox{0.85}{0.156}}
    & \boldres{\scalebox{0.85}{0.203}}
    & \secondres{\scalebox{0.85}{0.163}}
    & \secondres{\scalebox{0.85}{0.209}}
    & \scalebox{0.85}{0.174}
    & \scalebox{0.85}{0.214}
    & \scalebox{0.85}{0.186}
    & \scalebox{0.85}{0.227}
    & \scalebox{0.85}{0.195}
    & \scalebox{0.85}{0.271}
    & \scalebox{0.85}{0.202}
    & \scalebox{0.85}{0.261}
    & \scalebox{0.85}{0.172}
    & \scalebox{0.85}{0.220}
    & \scalebox{0.85}{0.195}
    & \scalebox{0.85}{0.252}
    & \scalebox{0.85}{0.221}
    & \scalebox{0.85}{0.306}
    & \scalebox{0.85}{0.217}
    & \scalebox{0.85}{0.296}
    & \scalebox{0.85}{0.173}
    & \scalebox{0.85}{0.223}
    & \scalebox{0.85}{0.266}
    & \scalebox{0.85}{0.336}
    \\

    & \scalebox{0.85}{192}
    & \boldres{\scalebox{0.85}{0.203}}
    & \boldres{\scalebox{0.85}{0.248}}
    & \secondres{\scalebox{0.85}{0.208}}
    & \secondres{\scalebox{0.85}{0.250}}
    & \scalebox{0.85}{0.221}
    & \scalebox{0.85}{0.254}
    & \scalebox{0.85}{0.234}
    & \scalebox{0.85}{0.265}
    & \scalebox{0.85}{0.209}
    & \scalebox{0.85}{0.277}
    & \scalebox{0.85}{0.242}
    & \scalebox{0.85}{0.298}
    & \scalebox{0.85}{0.219}
    & \scalebox{0.85}{0.261}
    & \scalebox{0.85}{0.237}
    & \scalebox{0.85}{0.295}
    & \scalebox{0.85}{0.261}
    & \scalebox{0.85}{0.340}
    & \scalebox{0.85}{0.276}
    & \scalebox{0.85}{0.336}
    & \scalebox{0.85}{0.245}
    & \scalebox{0.85}{0.285}
    & \scalebox{0.85}{0.307}
    & \scalebox{0.85}{0.367}
    \\

    & \scalebox{0.85}{336}
    & \secondres{\scalebox{0.85}{0.260}}
    & \secondres{\scalebox{0.85}{0.291}}
    & \boldres{\scalebox{0.85}{0.251}}
    & \boldres{\scalebox{0.85}{0.287}}
    & \scalebox{0.85}{0.278}
    & \scalebox{0.85}{0.296}
    & \scalebox{0.85}{0.284}
    & \scalebox{0.85}{0.301}
    & \scalebox{0.85}{0.273}
    & \scalebox{0.85}{0.332}
    & \scalebox{0.85}{0.287}
    & \scalebox{0.85}{0.335}
    & \scalebox{0.85}{0.280}
    & \scalebox{0.85}{0.306}
    & \scalebox{0.85}{0.282}
    & \scalebox{0.85}{0.331}
    & \scalebox{0.85}{0.309}
    & \scalebox{0.85}{0.378}
    & \scalebox{0.85}{0.339}
    & \scalebox{0.85}{0.380}
    & \scalebox{0.85}{0.321}
    & \scalebox{0.85}{0.338}
    & \scalebox{0.85}{0.359}
    & \scalebox{0.85}{0.395}
    \\

    & \scalebox{0.85}{720}
    & \boldres{\scalebox{0.85}{0.338}}
    & \secondres{\scalebox{0.85}{0.342}}
    & \secondres{\scalebox{0.85}{0.339}}
    & \boldres{\scalebox{0.85}{0.341}}
    & \scalebox{0.85}{0.358}
    & \scalebox{0.85}{0.347}
    & \scalebox{0.85}{0.356}
    & \scalebox{0.85}{0.349}
    & \scalebox{0.85}{0.379}
    & \scalebox{0.85}{0.401}
    & \scalebox{0.85}{0.351}
    & \scalebox{0.85}{0.386}
    & \scalebox{0.85}{0.365}
    & \scalebox{0.85}{0.359}
    & \scalebox{0.85}{0.345}
    & \scalebox{0.85}{0.382}
    & \scalebox{0.85}{0.377}
    & \scalebox{0.85}{0.427}
    & \scalebox{0.85}{0.403}
    & \scalebox{0.85}{0.428}
    & \scalebox{0.85}{0.414}
    & \scalebox{0.85}{0.410}
    & \scalebox{0.85}{0.419}
    & \scalebox{0.85}{0.428}
    \\

    \cmidrule(lr){2-26}

    & \scalebox{0.85}{Avg}
    & \boldres{\scalebox{0.85}{0.239}}
    & \boldres{\scalebox{0.85}{0.271}}
    & \secondres{\scalebox{0.85}{0.240}}
    & \boldres{\scalebox{0.85}{0.271}}
    & \scalebox{0.85}{0.258}
    & \secondres{\scalebox{0.85}{0.278}}
    & \scalebox{0.85}{0.265}
    & \scalebox{0.85}{0.285}
    & \scalebox{0.85}{0.264}
    & \scalebox{0.85}{0.320}
    & \scalebox{0.85}{0.271}
    & \scalebox{0.85}{0.320}
    & \scalebox{0.85}{0.259}
    & \scalebox{0.85}{0.287}
    & \scalebox{0.85}{0.265}
    & \scalebox{0.85}{0.315}
    & \scalebox{0.85}{0.292}
    & \scalebox{0.85}{0.363}
    & \scalebox{0.85}{0.309}
    & \scalebox{0.85}{0.360}
    & \scalebox{0.85}{0.288}
    & \scalebox{0.85}{0.314}
    & \scalebox{0.85}{0.338}
    & \scalebox{0.85}{0.382}
    \\
    \midrule
    \multirow{5}{*}{\rotatebox{90}{\scalebox{0.95}{Traffic}}} 
    & \scalebox{0.85}{96}
    & \boldres{\scalebox{0.85}{0.392}}
    & \boldres{\scalebox{0.85}{0.263}}
    & \scalebox{0.85}{0.462}
    & \scalebox{0.85}{0.285}
    & \secondres{\scalebox{0.85}{0.395}}
    & \secondres{\scalebox{0.85}{0.268}}
    & \scalebox{0.85}{0.526}
    & \scalebox{0.85}{0.347}
    & \scalebox{0.85}{0.644}
    & \scalebox{0.85}{0.429}
    & \scalebox{0.85}{0.805}
    & \scalebox{0.85}{0.493}
    & \scalebox{0.85}{0.593}
    & \scalebox{0.85}{0.321}
    & \scalebox{0.85}{0.650}
    & \scalebox{0.85}{0.396}
    & \scalebox{0.85}{0.788}
    & \scalebox{0.85}{0.499}
    & \scalebox{0.85}{0.587}
    & \scalebox{0.85}{0.366}
    & \scalebox{0.85}{0.612}
    & \scalebox{0.85}{0.338}
    & \scalebox{0.85}{0.613}
    & \scalebox{0.85}{0.388}
    \\

    & \scalebox{0.85}{192}
    & \secondres{\scalebox{0.85}{0.422}}
    & \secondres{\scalebox{0.85}{0.288}}
    & \scalebox{0.85}{0.473}
    & \scalebox{0.85}{0.296}
    & \boldres{\scalebox{0.85}{0.417}}
    & \boldres{\scalebox{0.85}{0.276}}
    & \scalebox{0.85}{0.522}
    & \scalebox{0.85}{0.332}
    & \scalebox{0.85}{0.665}
    & \scalebox{0.85}{0.431}
    & \scalebox{0.85}{0.756}
    & \scalebox{0.85}{0.474}
    & \scalebox{0.85}{0.617}
    & \scalebox{0.85}{0.336}
    & \scalebox{0.85}{0.598}
    & \scalebox{0.85}{0.370}
    & \scalebox{0.85}{0.789}
    & \scalebox{0.85}{0.505}
    & \scalebox{0.85}{0.604}
    & \scalebox{0.85}{0.373}
    & \scalebox{0.85}{0.613}
    & \scalebox{0.85}{0.340}
    & \scalebox{0.85}{0.616}
    & \scalebox{0.85}{0.382}
    \\

    & \scalebox{0.85}{336}
    & \boldres{\scalebox{0.85}{0.428}}
    & \boldres{\scalebox{0.85}{0.273}}
    & \scalebox{0.85}{0.498}
    & \scalebox{0.85}{0.296}
    & \secondres{\scalebox{0.85}{0.433}}
    & \secondres{\scalebox{0.85}{0.283}}
    & \scalebox{0.85}{0.517}
    & \scalebox{0.85}{0.334}
    & \scalebox{0.85}{0.674}
    & \scalebox{0.85}{0.420}
    & \scalebox{0.85}{0.762}
    & \scalebox{0.85}{0.477}
    & \scalebox{0.85}{0.629}
    & \scalebox{0.85}{0.336}
    & \scalebox{0.85}{0.605}
    & \scalebox{0.85}{0.373}
    & \scalebox{0.85}{0.797}
    & \scalebox{0.85}{0.508}
    & \scalebox{0.85}{0.621}
    & \scalebox{0.85}{0.383}
    & \scalebox{0.85}{0.618}
    & \scalebox{0.85}{0.328}
    & \scalebox{0.85}{0.622}
    & \scalebox{0.85}{0.337}
    \\

    & \scalebox{0.85}{720}
    & \boldres{\scalebox{0.85}{0.451}}
    & \boldres{\scalebox{0.85}{0.292}}
    & \scalebox{0.85}{0.506}
    & \scalebox{0.85}{0.313}
    & \secondres{\scalebox{0.85}{0.467}}
    & \secondres{\scalebox{0.85}{0.302}}
    & \scalebox{0.85}{0.552}
    & \scalebox{0.85}{0.352}
    & \scalebox{0.85}{0.683}
    & \scalebox{0.85}{0.424}
    & \scalebox{0.85}{0.719}
    & \scalebox{0.85}{0.449}
    & \scalebox{0.85}{0.640}
    & \scalebox{0.85}{0.350}
    & \scalebox{0.85}{0.645}
    & \scalebox{0.85}{0.394}
    & \scalebox{0.85}{0.841}
    & \scalebox{0.85}{0.523}
    & \scalebox{0.85}{0.626}
    & \scalebox{0.85}{0.382}
    & \scalebox{0.85}{0.653}
    & \scalebox{0.85}{0.355}
    & \scalebox{0.85}{0.660}
    & \scalebox{0.85}{0.408}
    \\

    \cmidrule(lr){2-26}

    & \scalebox{0.85}{Avg}
    & \boldres{\scalebox{0.8}{0.423}}
    & \boldres{\scalebox{0.8}{0.279}}
    & \scalebox{0.85}{0.484}
    & \scalebox{0.85}{0.297}
    & \secondres{\scalebox{0.85}{0.428}}
    & \secondres{\scalebox{0.85}{0.282}}
    & \scalebox{0.85}{0.529}
    & \scalebox{0.85}{0.341}
    & \scalebox{0.85}{0.667}
    & \scalebox{0.85}{0.426}
    & \scalebox{0.85}{0.760}
    & \scalebox{0.85}{0.473}
    & \scalebox{0.85}{0.620}
    & \scalebox{0.85}{0.336}
    & \scalebox{0.85}{0.625}
    & \scalebox{0.85}{0.383}
    & \scalebox{0.85}{0.804}
    & \scalebox{0.85}{0.509}
    & \scalebox{0.85}{0.610}
    & \scalebox{0.85}{0.376}
    & \scalebox{0.85}{0.624}
    & \scalebox{0.85}{0.340}
    & \scalebox{0.85}{0.628}
    & \scalebox{0.85}{0.379}
    \\

\bottomrule
  \end{tabular}
    \end{small}
  \end{threeparttable}
  }
\end{table}

On Exchange and Traffic, DST-Net leads in both average metrics, which is
consistent with its emphasis on cross-variate trend dependencies. It also
remains competitive on the ETT series and Weather, suggesting that the
model handles diverse temporal patterns. TimeMixer retains an advantage
on ETTm2, illustrating that the relative benefits vary across datasets.

Overall, these results suggest that adapting the modeling strategy to
each decomposed component is beneficial for long-term forecasting.
The Latent Graph Block targets cross-variate trend dependencies, while
Gaussian Multi-scale Attention models seasonal evolution at multiple
temporal scales. The following ablation studies further examine how
these two branches contribute to the observed gains.

\subsection{Short-term Forecasting}

We evaluate DST-Net on PEMS04 for short-term traffic forecasting, using
96 historical observations to predict the next 12 time steps. As shown
in Table~\ref{PEMS_results}, DST-Net achieves the lowest MAPE, RMSE, and
MAE among the compared methods.

Compared with the reproduced TimeMixer baseline, DST-Net reduces MAPE,
RMSE, and MAE by 12.5\%, 2.5\%, and 1.0\%, respectively. The larger gain
in MAPE indicates that the improvement is particularly evident in errors
measured relative to ground-truth traffic flow, alongside reductions in
absolute forecasting errors.

These results support the effectiveness of component-specific modeling
for short-term forecasting. In PEMS04, traffic states across sensors are
influenced by shared road conditions, making cross-sensor dependencies
relevant to prediction. The Latent Graph Block dynamically aggregates
information from relevant sensors to model correlated trend variations,
while Gaussian Multi-scale Attention models local fluctuations and
recurring temporal patterns at different granularities. Their
complementary roles allow DST-Net to capture both cross-sensor
co-movements and short-term temporal dynamics, consistent with its
effectiveness in both long-term and short-term forecasting.

\begin{table*}[t]
  \caption{Short-term forecasting results on the PEMS04 dataset. All input lengths are 96 and prediction lengths are 12. The results of TimeMixer are reproduced under the same experimental setting for a fair comparison.}
  \label{PEMS_results}
  \vspace{3pt}
  \centering
  \resizebox{\textwidth}{!}{
  \begin{threeparttable}
  \begin{small}
  \renewcommand{\multirowsetup}{\centering}
  \setlength{\tabcolsep}{0.7pt}
  \begin{tabular}{c|c|c|c|c|c|c|c|c|c|c|c|c|c|c}
    \toprule
    \multicolumn{2}{c}{\multirow{1}{*}{Models}} &
    \multicolumn{1}{c}{\rotatebox{0}{\scalebox{0.95}{\textbf{DST-Net (Ours)}}}}&
    \multicolumn{1}{c}{\rotatebox{0}{\scalebox{0.95}{TimeMixer}}} &
    \multicolumn{1}{c}{\rotatebox{0}{\scalebox{0.95}{iTransformer}}} &
    \multicolumn{1}{c}{\rotatebox{0}{\scalebox{0.95}{TiDE}}} &
    \multicolumn{1}{c}{\rotatebox{0}{\scalebox{0.95}{SCINet}}} &
    \multicolumn{1}{c}{\rotatebox{0}{\scalebox{0.95}{Crossformer}}} &
    \multicolumn{1}{c}{\rotatebox{0}{\scalebox{0.95}{PatchTST}}} &
    \multicolumn{1}{c}{\rotatebox{0}{\scalebox{0.95}{TimesNet}}} &
    \multicolumn{1}{c}{\rotatebox{0}{\scalebox{0.95}{MICN}}} &
    \multicolumn{1}{c}{\rotatebox{0}{\scalebox{0.95}{DLinear}}} &
    \multicolumn{1}{c}{\rotatebox{0}{\scalebox{0.95}{FEDformer}}} & 
    \multicolumn{1}{c}{\rotatebox{0}{\scalebox{0.95}{Stationary}}} & 
    \multicolumn{1}{c}{\rotatebox{0}{\scalebox{0.95}{Autoformer}}}
    \\
    \toprule

    \multirow{3}{*}{\scalebox{1.0}{\rotatebox{0}{Metric}}} 
    & \scalebox{1.0}{MAPE} & \boldres{\scalebox{1.0}{12.20}}& {\scalebox{1.0}{13.94}} & \scalebox{1.0}{13.42}& \scalebox{1.0}{14.76}& \secondres{\scalebox{1.0}{12.84}} & \secondres{\scalebox{1.0}{12.84}} & \scalebox{1.0}{16.65} &
    \scalebox{1.0}{13.15}  & \scalebox{1.0}{13.53} & \scalebox{1.0}{16.12} &
    \scalebox{1.0}{16.76}  & \scalebox{1.0}{14.85} &
    \scalebox{1.0}{16.70}
    \\
    & \scalebox{1.0}{RMSE} & \boldres{\scalebox{1.0}{ 31.97} }&{\scalebox{1.0}{ 32.80} }& \scalebox{1.0}{33.91}& \scalebox{1.0}{37.30}& \secondres{\scalebox{1.0}{32.31}} & \scalebox{1.0}{32.41} & \scalebox{1.0}{40.46} &
    \scalebox{1.0}{34.90}  & \scalebox{1.0}{34.39} & \scalebox{1.0}{39.51} &
    \scalebox{1.0}{41.81}  & \scalebox{1.0}{35.47} &
    \scalebox{1.0}{38.02}
    \\
    
    & \scalebox{1.0}{MAE} & \boldres{\scalebox{1.0}{19.81}} &\secondres{\scalebox{1.0}{20.00}} & \scalebox{1.0}{21.81}& \scalebox{1.0}{23.99}& {\scalebox{1.0}{20.35}} & \scalebox{1.0}{20.38} & \scalebox{1.0}{24.86} &
    \scalebox{1.0}{21.63}  & \scalebox{1.0}{21.62} & \scalebox{1.0}{24.62} &
    \scalebox{1.0}{26.51}  & \scalebox{1.0}{22.34} &
    \scalebox{1.0}{25.00}
    \\

    \bottomrule
  \end{tabular}
    \end{small}
  \end{threeparttable}
  }
\end{table*}

\subsection{Effectiveness of Different Components}

\begin{table}[t]
\caption{Component ablation results of DST-Net on Exchange, ETTh1, and Weather in terms of MSE.}
\label{tab:ablation_results}
\centering
\resizebox{\linewidth}{!}{
\begin{small}
\begin{tabular}{l|ccccc|ccccc|ccccc}
\toprule
Datasets
& \multicolumn{5}{c|}{Exchange}
& \multicolumn{5}{c|}{ETTh1}
& \multicolumn{5}{c}{Weather} \\
\midrule
Prediction Length $T$
& 96 & 192 & 336 & 720 & Avg
& 96 & 192 & 336 & 720 & Avg
& 96 & 192 & 336 & 720 & Avg \\
\midrule

w/o Dec
& 0.091 & 0.183 & 0.341 & 0.823 & 0.360
& 0.393 & 0.434 & 0.459 & 0.467 & 0.438
& 0.171 & 0.221 & 0.278 & 0.361 & 0.258 \\

w/o LG (Trend)
& 0.094 & 0.187 & 0.348 & 0.842 & 0.368
& 0.383 & 0.424 & 0.449 & 0.455 & 0.428
& 0.160 & 0.209 & 0.266 & 0.346 & 0.245 \\

w/o GMA (Season)
& 0.086 & 0.176 & 0.329 & 0.806 & 0.349
& 0.390 & 0.432 & 0.456 & 0.463 & 0.435
& 0.175 & 0.225 & 0.283 & 0.369 & 0.263 \\
\midrule

\textbf{DST-Net (Ours)}
& \textbf{0.082} & \textbf{0.168} & \textbf{0.320}
& \textbf{0.785} & \textbf{0.339}
& \textbf{0.375} & \textbf{0.415} & \textbf{0.439}
& \textbf{0.444} & \textbf{0.418}
& \textbf{0.156} & \textbf{0.203} & \textbf{0.260}
& \textbf{0.338} & \textbf{0.239} \\

\bottomrule
\end{tabular}
\end{small}
}
\end{table}

To evaluate the contribution of each component, we construct three variants of DST-Net: \textit{w/o Dec} bypasses series decomposition, \textit{w/o LG} removes Latent Graph Block from the trend branch, and \textit{w/o GMA} replaces Gaussian Multi-scale Attention with vanilla self-attention. As shown in Table~\ref{tab:ablation_results}, removing decomposition consistently increases the average MSE, demonstrating the importance of separating heterogeneous temporal components. Removing the Latent Graph Block causes the largest degradation on Exchange, increasing the average MSE from 0.339 to 0.368, which verifies its effectiveness in capturing cross-variate trend dependencies. In contrast, removing Gaussian Multi-scale Attention has the greatest effect on Weather, where the average MSE increases from 0.239 to 0.263. This confirms the benefit of Gaussian locality constraints for modeling complex seasonal variations. Overall, the results demonstrate the complementary contributions of decomposition, latent graph modeling, and Gaussian multi-scale temporal modeling.

\subsection{Effectiveness of Component-specific Modeling}

\begin{table*}[t]
\caption{Effectiveness of component-specific modeling}
\label{tab:assignment_ablation}
\centering
\resizebox{\textwidth}{!}{
\begin{tabular}{l|ccccc|ccccc|ccccc}
\toprule
Datasets
& \multicolumn{5}{c|}{Exchange}
& \multicolumn{5}{c|}{ETTh1}
& \multicolumn{5}{c}{Weather} \\
\midrule
Prediction Length
& 96 & 192 & 336 & 720 & Avg
& 96 & 192 & 336 & 720 & Avg
& 96 & 192 & 336 & 720 & Avg \\
\midrule

Both GMA
& 0.089 & 0.180 & 0.335 & 0.817 & 0.355
& 0.385 & 0.427 & 0.451 & 0.457 & 0.430
& 0.164 & 0.211 & 0.268 & 0.348 & 0.248 \\

Both LG
& 0.087 & 0.177 & 0.333 & 0.812 & 0.352
& 0.388 & 0.430 & 0.454 & 0.461 & 0.433
& 0.174 & 0.223 & 0.280 & 0.362 & 0.260 \\

Swapped
& 0.096 & 0.189 & 0.342 & 0.841 & 0.367
& 0.397 & 0.439 & 0.464 & 0.472 & 0.443
& 0.179 & 0.229 & 0.287 & 0.371 & 0.267 \\
\midrule

\textbf{DST-Net (Ours)}
& \textbf{0.082} & \textbf{0.168} & \textbf{0.320}
& \textbf{0.785} & \textbf{0.339}
& \textbf{0.375} & \textbf{0.415} & \textbf{0.439}
& \textbf{0.444} & \textbf{0.418}
& \textbf{0.156} & \textbf{0.203} & \textbf{0.260}
& \textbf{0.338} & \textbf{0.239} \\

\bottomrule
\end{tabular}
}
\end{table*}

To verify whether the branch assignment is essential, we compare DST-Net with three alternative designs: \textit{Both GMA}, which applies GMA to both components; \textit{Both LG}, which applies LG to both components; and \textit{Swapped}, which exchanges the modeling strategies of the seasonal and trend branches. As shown in Table~\ref{tab:assignment_ablation}, both homogeneous variants underperform the complete model, indicating that applying an identical modeling strategy overlooks the distinct properties of the decomposed components. Moreover, \textit{Swapped} produces the largest degradation, increasing the average MSE from 0.339/0.418/0.239 to 0.367/0.443/0.267 on Exchange, ETTh1, and Weather, respectively. These results confirm that the performance gain arises not only from the individual modules, but also from assigning GMA to seasonal evolution and LG to cross-variate trend modeling.

\subsection{Effectiveness of Different Gaussian Scales in GMA}

\begin{table}[t]
\caption{Impact of different Gaussian scales in GMA on the Weather dataset.}
\label{tab:gma_scales}
\centering
\resizebox{0.6\linewidth}{!}{
\begin{tabular}{l|cccc|c}
\toprule
Variant & Narrow & Medium & Wide & Global & MSE \\ \midrule
Single Scale & \checkmark & - & - & - & 0.252 \\
Single Scale & - & \checkmark & - & - & 0.249 \\
Single Scale & - & - & \checkmark & - & 0.255 \\
Single Scale & - & - & - & \checkmark & 0.263 \\ 
\midrule
\textbf{Multi-scale} & \checkmark & \checkmark & \checkmark & - & 0.243 \\
\textbf{Multi-scale} & \checkmark & \checkmark & \checkmark & \checkmark & \textbf{0.239} \\
\bottomrule
\end{tabular}
}
\vspace{3pt}
\end{table}

To evaluate the impact of the design of our Gaussian Multi-scale Attention (GMA) mechanism, we conduct a detailed analysis on the Weather dataset by varying the combination of Gaussian scales. As shown in Table~\ref{tab:gma_scales}, while single-scale variants such as the Narrow or Medium capture specific temporal granularities, they yield suboptimal results.  Our full multi-scale configuration, which fuses Narrow, Medium, Wide, and Global perspectives, achieves the superior MSE of 0.239. This improvement validates that the seasonal component's features are best captured through a hierarchical focus, where learnable Gaussian priors across multiple scales allow the model to simultaneously filter high-frequency noise and retain long-term periodic dependencies, thereby ensuring a more robust representation of complex seasonal patterns.

\subsection{Qualitative Analysis: Component Disentanglement}

\begin{figure}[t]
    \centering
    \includegraphics[width=0.5\linewidth]{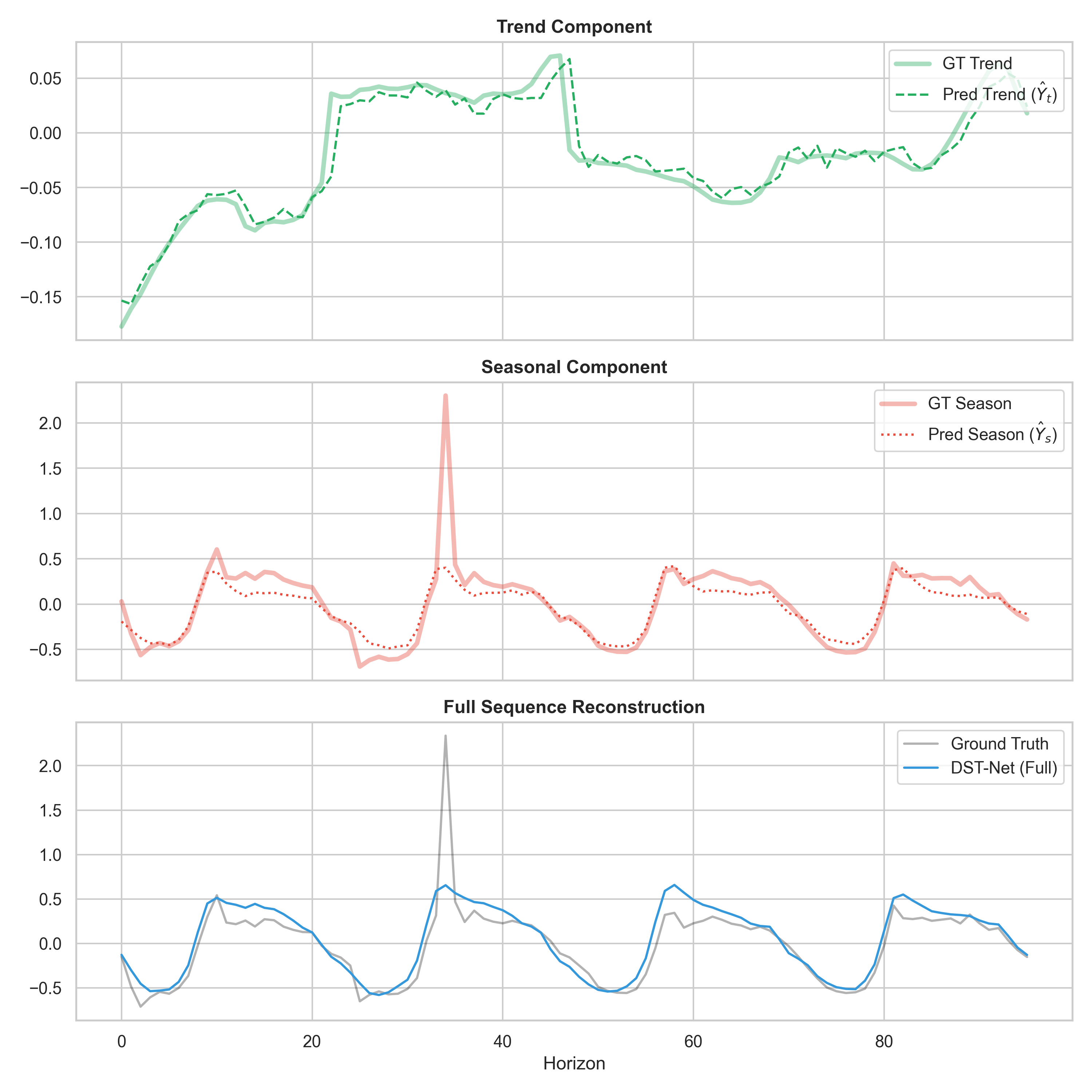}
    \caption{Disentanglement analysis on the Traffic dataset. We compare the ground truth (GT) decomposition with the predictions of our trend branch ($\hat{Y}_t$), seasonal branch ($\hat{Y}_s$), and full sequence.}
    \label{disentanglement}
\end{figure}

Figure~\ref{disentanglement} illustrates the complementary behavior of the two forecasting branches. The trend prediction follows the smooth evolution of the decomposed target, while the seasonal prediction captures the recurring oscillatory structure. Nevertheless, the seasonal branch underestimates the isolated abrupt peak, indicating that rare high-amplitude events remain challenging. The reconstructed sequence follows the overall temporal pattern, suggesting that the two branches provide complementary information for forecasting. 

\FloatBarrier
\section{Conclusion}

In this paper, we propose DST-Net, a novel decomposed spatio-temporal network designed to address the inherent inductive bias conflicts between seasonal and trend components in time series forecasting. By identifying that the trend component is dominated by global spatial coupling while the seasonal component relies on local temporal evolution, we move beyond traditional homogeneous modeling strategies. Our architecture integrates a Latent Graph Block to capture system-wide co-movements within the trend branch and a Gaussian Multi-scale Attention mechanism to mitigate attention distraction in the seasonal branch. Extensive experiments on real-world benchmarks demonstrate that DST-Net consistently outperforms leading models in both long-term and short-term forecasting tasks. 

\section{Acknowledgments}
This work is supported in part by the Key Research Project of Pengcheng Laboratory under Grant PCL2025A14.

\bibliographystyle{splncs04}
\bibliography{main}
\end{document}